\documentclass[11pt]{article}
\usepackage{arabtex}
\usepackage{emnlp2021}
\usepackage{times,latexsym}
\usepackage{url}
\usepackage[T1]{fontenc}
\usepackage[utf8]{inputenc}
\usepackage{utf8}

\usepackage{pgf}
\usepackage{tikz-dependency}

\usepackage[export]{adjustbox}

\usepackage{verbatim}
\usepackage{graphicx,wrapfig,lipsum}

\usepackage{multirow}
\usepackage{amssymb}
\usepackage{amsmath}
\usepackage{algorithmicx}
\usepackage[ruled]{algorithm}
\usepackage[noend]{algpseudocode}
\usepackage{latexsym}
\usepackage{textcomp}
\usepackage{tikz}
\usepackage{xcolor}
\usepackage{pgfplots}
\usetikzlibrary{patterns}
\usepackage{subcaption}

\newcommand{\printfnsymbol}[1]{%
 	\textsuperscript{\@fnsymbol{#1}}%
 }
\usepackage{microtype}
\usepackage{pifont}
\newcommand{\cmark}{\ding{51}}%
\newcommand{\xmark}{\ding{55}}%

\title{``Wikily'' Supervised Neural Translation Tailored to Cross-Lingual Tasks}

\author{Mohammad Sadegh Rasooli$^{1}$\thanks{~~Research was conducted at The University of Pennsylvania.}\ \ Chris Callison-Burch$^{2}$\ \  Derry Tanti Wijaya$^{3}$ \\
	\normalsize{~$^1$Microsoft}\\[-.05cm]
	\normalsize{~$^2$Department of Computer and Information Science,  University of Pennsylvania}\\[-.05cm]
	\normalsize{~$^3$Department of Computer Science,  Boston University}\\[-.05cm]
	{\small\tt mrasooli@microsoft.com,\ ccb@seas.upenn.edu,\  wijaya@bu.edu}  \\
}

\date{}

\begin{document}
\maketitle
\begin{abstract}
We present a simple but effective approach for leveraging Wikipedia for neural machine translation as well as cross-lingual tasks of image captioning and dependency parsing without using any direct supervision from external parallel data or supervised models in the target language. We show that first sentences and titles of linked Wikipedia pages, as well as cross-lingual image captions, are strong signals for a seed parallel data  to extract bilingual dictionaries and cross-lingual word embeddings for mining parallel text from Wikipedia. Our final model achieves high BLEU scores that are close to or sometimes higher than strong \emph{supervised} baselines in low-resource languages; e.g. supervised BLEU of 4.0 versus 12.1 from our model in English-to-Kazakh.  Moreover, we tailor our \emph{``wikily'' supervised} translation models to unsupervised image captioning, and cross-lingual dependency parser transfer. In image captioning, we train a multi-tasking machine translation and image captioning pipeline for Arabic and English from which the Arabic training data is a translated  version of the English captioning data, using our wikily-supervised translation models. Our captioning results on Arabic are slightly \emph{better} than that of its supervised model. In dependency parsing, we translate a large amount of monolingual text, and use it as artificial training data in an \emph{annotation projection} framework. We show that our model outperforms recent work on cross-lingual transfer of dependency parsers.
\end{abstract}

\section{Introduction}

Developing machine translation models without using bilingual parallel text is an intriguing research problem with real applications: obtaining a large volume of parallel text for many languages is hard if not impossible. Moreover, translation models could be used in downstream cross-lingual tasks in which annotated data does not exist for some languages. There has recently been a great deal of interest in unsupervised neural machine translation~(e.g.~\newcite{artetxe-etal-2018-unsupervised,lample2018unsupervised,lample-etal-2018-phrase,xlm_paper,song2019mass,kim-etal-2020-unsupervised,tae2020meta}). Unsupervised neural machine translation models often perform nearly as well as supervised models when translating between similar languages, but they fail to perform well in low-resource or distant languages~\cite{kim-etal-2020-unsupervised} or out-of-domain monolingual data~\cite{marchisio-etal-2020-unsupervised}. In practice, the highest need for unsupervised models is to expand beyond high resource, similar European language pairs. 

There are two key goals in this paper: Our first goal is developing accurate translation models for low-resource distant languages \emph{without} any supervision from a supervised model or gold-standard parallel data. Our second goal is to show that our machine translation models can be directly tailored to downstream natural language processing tasks. In this paper, we showcase our claim in cross-lingual image captioning and cross-lingual transfer of dependency parsers, but this idea is applicable to a wide variety of tasks.

\begin{figure}[t!]
\centering
\includegraphics[width=0.45\textwidth]{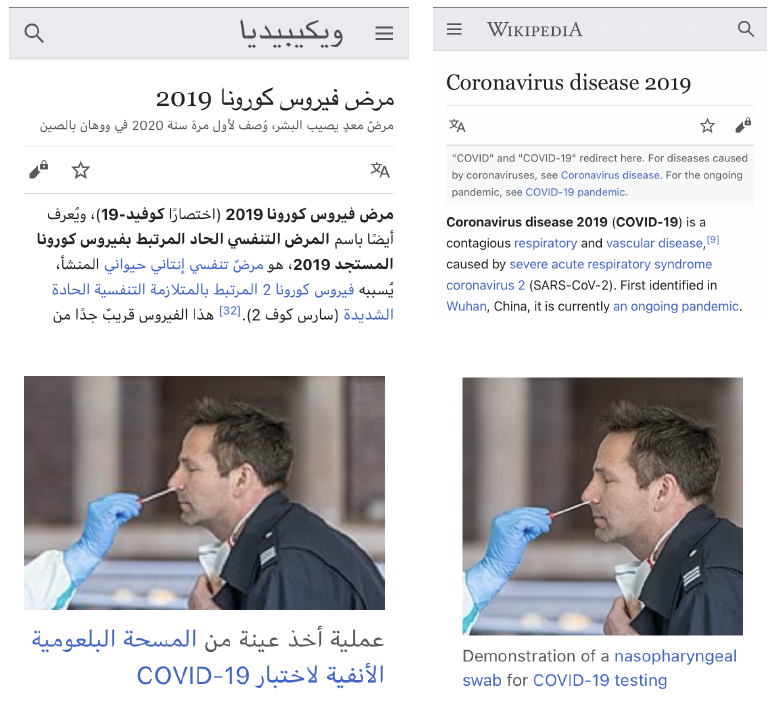}
 
  
  
  
\caption{A pair of Wikipedia documents in Arabic and English, along with a same image with two captions.}
\label{fig:wiki_image}
\vspace{-1em}
\end{figure}

We present a fast and accurate approach for learning translation models using Wikipedia. Unlike \emph{unsupervised machine translation} that solely relies on raw monolingual data, we believe that we should not neglect the availability of incidental supervisions from online resources such as Wikipedia. Wikipedia contains articles in nearly 300 languages and more languages might be added in the future, including indigenous languages and dialects of different regions in the world. Different from similar recent work~\cite{schwenk2019wikimatrix}, we do not rely on any supervision from supervised translation models. Instead, we leverage the fact that many first sentences in linked Wikipedia pages are rough translations, and furthermore, many captions of the same images are similar sentences, sometimes translations. Figure~\ref{fig:wiki_image} shows a real example of a pair of linked Wikipedia pages in Arabic and English in which the titles, first sentences, and also the image captions are rough translations of each other. Our method learns a seed bilingual dictionary from a small collection of first sentence pairs, titles and captions, and then learns cross-lingual word embeddings. We make use of cross-lingual word embeddings to extract parallel sentences from Wikipedia. Our experiments show that our approach improves over strong unsupervised translation models for low-resource languages: we improve the BLEU score of English$\rightarrow$Gujarati from $0.6$ to $15.2$, and English$\rightarrow$Kazakh from $0.8$ to $12.1$. 

In the realm of downstream tasks, we show that we can easily use our translation models to generate high-quality translations of MS-COCO~\cite{chen2015microsoft} and Flickr~\cite{hodosh2013framing} datasets, and train a cross-lingual image captioning model in a multi-task pipeline paired with machine translation in which the model is initialized by the parameters from our translation model. Our results on Arabic captioning show a BLEU score of $5.72$ that is slightly better than a supervised captioning model with a BLEU score of $5.22$. As another task, in \emph{dependency parsing}, we first translate a large amount of monolingual data using our translation models and then apply transfer using the \emph{annotation projection} method~\cite{Yarowsky:2001:IMT:1072133.1072187,hwa2005bootstrapping}. Our results show that our approach performs similarly compared to using gold-standard parallel text in high-resource scenarios, and significantly better in low-resource languages. 


A summary of our contribution is as follows: 1) We propose a simple, fast and effective approach towards using the Wikipedia monolingual data for machine translation without any explicit supervision. Our mining algorithm easily scales on large comparable data using limited computational resources. We achieve very high BLEU scores for distant languages, especially those in which current unsupervised methods perform very poorly. 2) We propose novel methods for leveraging our current translation models in image captioning. We show that how a combination of translating caption training data, and multi-task learning with English captioning as well as translation improves the performance. Our results on Arabic shows results slightly superior to that of a supervised captioning model trained on gold-standard datasets. 3) We propose a novel modification to the annotation projection method to be able to leverage our translation models. Our results on dependency parsing performs better than previous work in most cases, and performs similarly to using gold-standard parallel datasets.

Our translation and captioning code and models are publicly available online\footnote{Our code: \url{https://github.com/rasoolims/ImageTranslate}. Our modification to Stanza for training on partially projected trees: \url{https://github.com/rasoolims/stanza}.}.
\section{Background}

\paragraph{Supervised neural machine translation}
 Supervised machine translation uses a parallel text ${\cal P}=\{(s_i,t_i)\}_{i=1}^{n}$ in which each sentence $s_i \in l_1$ is a translation of $t_i \in l_2$. \emph{Neural machine translation} uses sequence-to-sequence models with attention~\cite{cho-etal-2014-learning,Bahdanau2015NeuralMT,vaswani2017attention} for which the likelihood of training data is maximized by maximizing the log-likelihood of predicting each target word given its previous predicted words and source sequence:
\[
{\cal L}({\cal P}) = \sum_{i=1}^{n} \sum_{j=1}^{|t_i|} \log p(t_{i,j} | t_{i,k<j}, s_i; \theta)
\]
where $\theta$ is a collection of parameters to be learned. 

\paragraph{Unsupervised neural machine translation}
Unsupervised neural machine translation does not have access to any parallel data. Instead, it tailors monolingual datasets ${\cal M}_{l_1}$ and ${\cal M}_{l_2}$ for learning multilingual language models. These language models usually mask parts of every input sentence, and try to uncover the masked words~\cite{devlin-etal-2019-bert}. The monolingual language models are used along with iterative back-translation~\cite{hoang-etal-2018-iterative} to learn unsupervised translation. An input sentence $s$ is translated to $t'$ using current model $\theta$, then the model assumes that $(t', s)$ is a gold-standard translation, and uses the same training objective as of supervised translation. 

 \begin{figure*}
   \centering
\includegraphics[width=\textwidth]{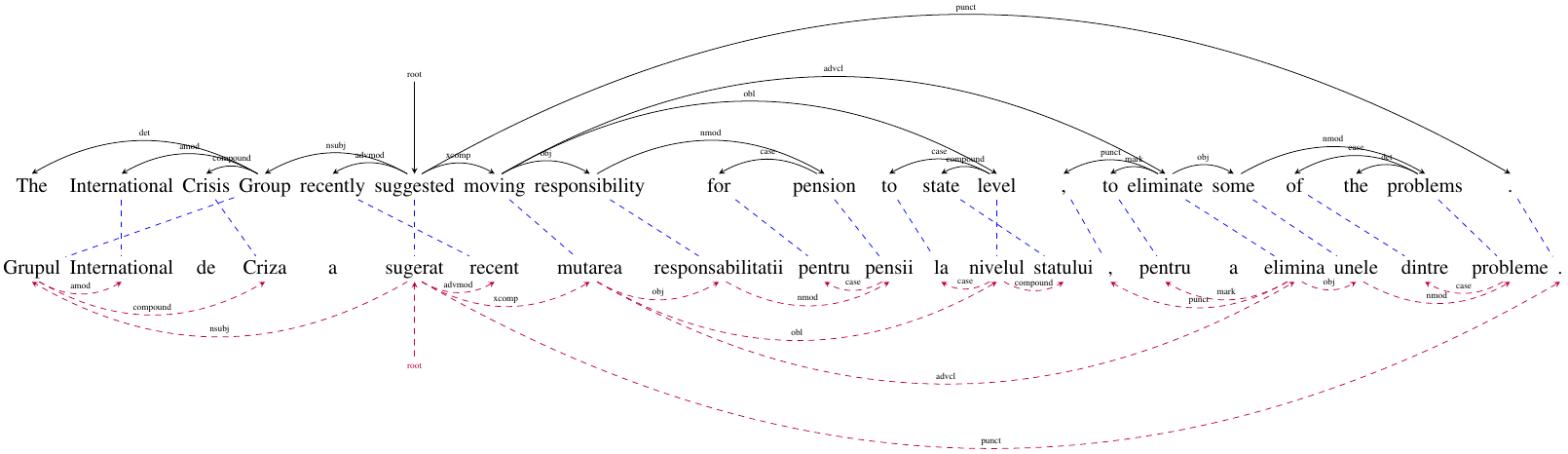}

\vspace{0.0cm}
   \caption[skip=0pt]{An example of annotation projection for which the source on top is a translation of the Romanian target via our \emph{wikily} translation model. The supervised source tree is projected using intersected word alignments. }
     \label{fig:annotation_projection}

 \end{figure*}

\paragraph{Dependency parsing}
Dependency parsing algorithms capture the best scoring dependency trees for sentences among an exponential number of possible dependency trees. A valid dependency tree for a sentence $s=s_1,\ldots,s_n$ assigns heads $h_i$ for each for word $s_i$ where $1 \leq i\leq n$, $0\leq h_i\leq n$ and $h_i \neq i$. The zeroth word represents a dummy root token as an indicator for the root of the sentence. For more details about efficient parsing algorithms, we encourage the reader to see \newcite{kubler2009dependency}.  

\paragraph{Annotation projection}
Annotation projection is an effective method for transferring supervised annotation from a rich-resource language to a low-resource language through translated text~\cite{Yarowsky:2001:IMT:1072133.1072187}. Having a parallel data ${\cal P} = \{(s_i, t_i)\}_{i=1}^{n}$, and supervised source annotations for source sentences $s_i$, we transfer those annotations through word translation links $0 \leq a_{i}^{(j)} \leq |t_i|$ for $1 \leq j\leq |s_i|$ where $ a_{i}^{(j)} =0$ shows a {\tt null} alignment. The alignment links are learned in an unsupervised fashion using unsupervised word alignment algorithms \cite{giza}. In dependency parsing, if $h_i = j$ and $a^{(j)} = k$ and $a^{(i)}=m$, we project a dependency $k\rightarrow m$ (i.e. $h_m=k$) to the target side. Previous work~\cite{rasooli-collins-2017-cross,rasooli-collins-2019-low} has shown that annotation projection only works when a large amount of translation data exists. In the absence of parallel data, we create artificial parallel data using our translation models. Figure~\ref{fig:annotation_projection} shows an example of annotation projection using translated text.

\newcommand{\docs}{w}
\newcommand{\captions}{{\cal C}}
\newcommand{\images}{{\cal I}}
\newcommand{\seed}{{\cal S}}
\newcommand{\comp}{\texttt{COMP}}
\newcommand{\first}{{\cal F}}
\newcommand{\dict}{{\cal D}}
\newcommand{\bitext}{{\cal P}}
\newcommand{\titles}{{\cal T}}
\newcommand{\embed}{v}

\section{Learning Translation from Wikipedia}
The key component of our approach is to leverage the multilingual cues from linked Wikipedia pages across languages. Wikipedia is a great comparable data in which many of its pages explain entities in the world in different languages. 
In most cases, first sentences define or introduce the mentioned entity in that page (e.g.  Figure~\ref{fig:wiki_image}). Therefore, we observe that many first sentence pairs in linked Wikipedia documents are rough translations of each other. Moreover, captions of images in different languages are usually similar but not necessarily direct translations of each other. We leverage this information to extract many parallel sentences from Wikipedia without using any external supervision. In this section, we describe our algorithm which is briefly shown in Figure~\ref{fig:alg_wiki}.

\subsection{Data Definitions}
For languages $e$ and $f$ in which $e$ is English and $f$ is a low-resource target language of interest, there are Wikipedia documents $\docs_e= \{\docs^{(e)}_1 \ldots \docs^{(e)}_n\}$ and  $\docs_f= \{\docs^{(f)}_1 \ldots \docs^{(f)}_m\}$. We refer to $\docs^{(l)}_{(i,j)}$ as the $j$th sentence in the $i$th document for language $l$.  A subset of these documents are aligned (using Wikipedia \emph{languages links}). Thus we have an aligned set of document pairs  in which we can easily extract many sentence pairs that are potentially translations of each other. A smaller subset $\first$ is the set of first sentences in Wikipedia $(\docs^{(e)}_{(i,1)},\docs^{(f)}_{(i',1)})$ in which documents $i$ and $i'$ are linked and their first sentence lengths are in a  similar range.  In addition to text content, Wikipedia has a large set of images. Each image comes along with one or more captions, sometimes in different languages. A small subset of these images have captions both in English and the target language. We refer to this set as $\captions$. We use the set of all caption pairs ($\captions$), title pairs ($\titles$), and first sentences ($\first$) as the seed parallel data: $\seed = \first \cup \captions \cup \titles$. 

\subsection{Bilingual Dictionary Extraction and Cross-Lingual Word Embeddings}
Having the seed parallel data $\seed$, we run unsupervised word alignment~\cite{dyer-etal-2013-simple} in both English-to-target, and target-to-English directions. We use the intersected alignments to extract highly confident word-to-word connections. Finally, we pick the most frequently aligned word for each word in English as translation. This set serves as a bilingual dictionary $\dict$.

Given two monolingual trained word embeddings $\embed_e \in \mathbb{R}^{N_e\times d}$ and $\embed_f \in \mathbb{R}^{N_f\times d}$, and the extracted bilingual dictionary $\dict$, we use the method of \newcite{faruqui-dyer-2014-improving} to project these two embedding vectors to a shared cross-lingual space.\footnote{There are more recent approaches such as \cite{lample2018word}. Comparing different embedding methods is not the focus of this paper, thereby we leave further investigation to future work.} This method uses a bilingual dictionary along with canonical correlation analysis (CCA) to learn two projection matrices to map each embedding vector to a shared space  $\embed'_e \in \mathbb{R}^{N_e\times d'}$ and $\embed'_f \in \mathbb{R}^{N_f\times d'}$ where $d'\leq d$. 

\input{alg}

\subsection{Mining Parallel Sentences} 
We use cross-lingual embedding vectors $\embed'_e \in \mathbb{R}^{N_e\times d}$ and $\embed'_f \in \mathbb{R}^{N_f\times d'}$ for calculating the cosine similarity between pairs of words. Moreover, we use the extracted bilingual dictionary to boost the accuracy of the scoring function.  For a pair of sentences $(s,t)$ where $s=s_1 \ldots s_n$ and $t=t_1 \ldots t_m$, after filtering sentence pairs with different numerical values (e.g. sentences containing $2019$ in the source and $1987$ in the target), we use a modified version of cosine similarity between words:
\[
    sim(s_i, t_j)= 
\begin{cases}
    1.0,& \text{if }  (s_i,t_j)\in \dict \\
    cos(s_i,t_j),              & \text{otherwise}
\end{cases}
\]
Using the above definition of word similarity, we use the average-maximum similarity between pairs of sentences. 
\[
    score(s, t) = \frac{\sum_{i=1}^{n}\max_{j=1}^{m} sim(s_i, t_i)}{n}
\]
From a pool of candidates, we pick  those pairs that have the highest score in both directions. 


\subsection{Leveraging Similar Languages}
In many low-resource scenarios, the number of paired documents is very small, leading to a small number and often noisy extracted parallel sentences. To alleviate this problem to some extent, we assume to have another language $g$ in which $g$ has a large lexical overlap with the target language $f$ (such as $g$=Russian and $f$=Kazakh). We assume that a parallel data exists between language $g$ and English, and we can use it both as an auxiliary parallel data in training, and also for extracting extra lexical entries for the bilingual dictionaries: as shown in Figure~\ref{fig:alg_wiki}, we supplement the extracted bilingual dictionary from seed parallel data with the bilingual dictionary extracted from related language parallel data.

\subsection{Translation Model}\label{sec:model}
We use a standard sequence-to-sequence transformer-based translation model~\cite{vaswani2017attention} with a six-layer BERT-based~\cite{devlin-etal-2019-bert} encoder-decoder architecture from HuggingFace~\cite{wolf2019huggingface} and Pytorch~\cite{paszke2019pytorch} with a shared SentencePiece~\cite{sentencepiece} vocabulary. All input and output token embeddings are summed up with the language id embedding. First tokens of every input and output sentence are shown by the language ID. Our training pipeline assumes that the encoder and decoder are shared across different languages, except that we use a separate output layer for each language in order to prevent input copying~\cite{artetxe2018unsupervised,sen-etal-2019-multilingual}. We pretrain the model on a tuple of three Wikipedia datasets for the three languages $g$, $f$, and $e$ using the MASS model~\cite{song2019mass}. The MASS model masks a contiguous span of input tokens, and recovers that span in the output sequence. 

To facilitate multi-task learning with image captioning, our model has an image encoder that is used in cases of image captioning (more details in \S\ref{sec:captioning}). In other words, the decoder is shared between the translation and captioning tasks. We use the pretrained ResNet-152 model~\cite{resnet} from Pytorch to encode every input image. We extract the final layer as a $7\times 7$ grid vector ($g \in \mathbb{R}^{7 \times 7 \times d_g}$), and project it to a new space by a linear transformation ($g' \in \mathbb{R}^{49\times d_t}$), and then add location embeddings ($l \in \mathbb{R}^{49 \times d_t}$) by using entry-wise addition. Afterwards, we assume that the $49$ vectors are encoded text representations as if a sentence with $49$ words occurs. This is similar to but not exactly the same as the Virtex model~\cite{virtex}.

\subsection{Back-Translation: One-shot and Iterative}
Finally, we use the back-translation technique to improve the quality of our models. Back-translation is done by translating a large amount of monolingual text to and from the target language. The translated texts serve as noisy input text along with the monolingual data as the silver-standard translations. Previous work~\cite{sennrich-etal-2016-improving,edunov-etal-2018-understanding} has shown that back-translation is a very simple but effective technique to improve the quality of translation models. Henceforth, we refer to this method as \emph{one-shot back-translation}. Another approach is to use \emph{iterative back-translation}~\cite{hoang-etal-2018-iterative}, the most popular approach in unsupervised translation~\cite{artetxe2018unsupervised,xlm_paper,song2019mass}. The main difference from one-shot translation is that the model uses an online approach, and updates its parameters in every batch. 

We empirically find \emph{one-shot back-translation} faster to train but with much less potential to reach a high translation accuracy. A simple and effective way to have both a reliable and accurate model is to first initialize a model with one-shot back-translation, and then apply iterative back-translation. The model that is initialized with a more accurate model reaches a higher accuracy.

\section{Cross-Lingual Tasks}
In this section, we describe our approaches for tailoring our translation models to cross-lingual tasks. Note that henceforth we assume that our translations model training is finished, and we have access to trained translation models for cross-lingual tasks. 

\subsection{Cross-Lingual Image Captioning}\label{sec:captioning}
\begin{figure}[t!]
    \centering
    \includegraphics[width=0.48\textwidth]{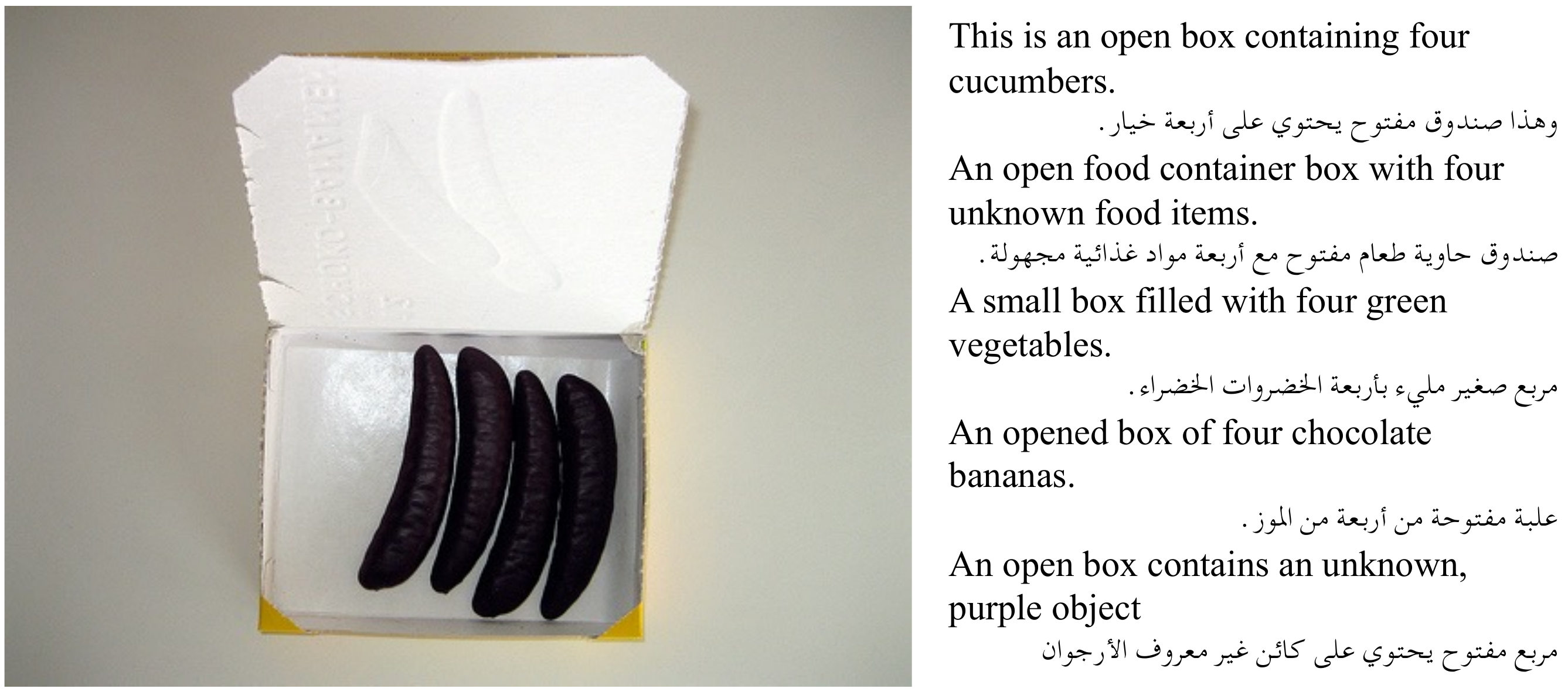}
    \caption{An image from MS-Coco~\protect\cite{chen2015microsoft} with gold-standard English captions, and Arabic translations from our \emph{wikily} translation model.}
    \label{fig:caption_example}
\end{figure}

Having gold-standard image captioning training data ${\cal I} = \{(I_i, c_i)\}_{i=1}^{n}$ where $I_i$ is the image as pixel values, and $c_i = c_i^{(1)}, \ldots, c_i^{k_i}$ as the textual description with $k_i$ words, our goal is to learn a captioning model that is able to describe new (unseen) images. As described in \S\ref{sec:model}, we use a transformer decoder from our translation model and a ResNet image encoder~\cite{resnet} for our image captioning pipeline.  Unfortunately, annotated image captioning datasets do not exist in many languages. Having our translation model parameter $\theta^*_{\rightleftarrows}$, we can use its translation functionality to translate each caption $c_i$ to $c'_i = translate(c_i | \theta^*_{\rightleftarrows})$. Afterwards, we will have a translated annotated dataset ${\cal I'} = \{(I_i, c'_i)\}_{i=1}^{n}$ in which the textual descriptions are not gold-standard but translations from the English captions. Figure~\ref{fig:caption_example} shows a real example from MS-Coco~\cite{chen2015microsoft} in which Arabic translations are provided by our translation model. Furthermore, to augment our learning capability, we initialize our decoder with decoding parameters of $\theta^*_{\rightleftarrows}$, and also continue training with both English captioning and translation.

\subsection{Cross-Lingual Dependency Parsing}
Assuming that we have a large body of monolingual text, we translate that monolingual text to create artificial parallel data. We run unsupervised word alignments on the artificial parallel text. Following previous work~\cite{rasooli-collins-2015-density,ma-xia:2014:P14-1}, we run Giza++~\cite{och2000giza} alignments on both source-to-target and target-to-source directions, and extract intersected alignments to keep high-precision one-to-one alignments. We run a supervised dependency parser of English as our rich-resource language. Then, we project dependencies to the target language sentences via word alignment links. Inspired by previous work~\cite{rasooli-collins-2015-density}, to remove noisy projections, we keep those sentences that at least $50\%$ of words or $5$ consecutive words in the target side have projected dependencies. 

\section{Experiments}\label{sec:expriments}
In this section, we provide details about our experimental settings and results for translation, captioning, and dependency parsing. We put more details about our settings as well as thorough analysis of our results in the supplementary material. 

\subsection{Datasets and Settings}
\paragraph{Languages} We focus on four language pairs: Arabic-English, Gujarati-English, Kazakh-English, and Romanian-English. We choose these pairs to provide enough evidence that our model works in distant languages, morphologically-rich languages, as well as similar languages.  As for similar languages, we use Persian for Arabic (written with very similar scripts and have many words in common), Hindi for Gujarati (similar languages), Russian for Kazakh (written with the same script), and Italian for Romanian (Romance languages).

\paragraph{Monolingual and Translation Datasets} 

We use a shared SentencePiece vocabulary~\cite{sentencepiece} with size 60K. Table~\ref{tab:par_sizes} shows the sizes of Wikipedia data in different languages. For evaluation, we use the Arabic-English UN data~\cite{ziemski2016united}, WMT 2019 data~\cite{barrault-etal-2019-findings} for Gujarati-English and Kazakh-English, and WMT 2016 shared task data~\cite{bojar-etal-2016-findings} for Romanian-English. Following previous work~\cite{sennrich-etal-2016-edinburgh}, diacritics are removed from the Romanian data. More details about other datasets and their sizes, we refer the reader to the supplementary material.

\begin{table}[]
    \centering \tiny
        \setlength{\tabcolsep}{5pt}
    \begin{tabular}{l|c c c c}
    \hline\hline
        Direction  & ar$\rightleftarrows$en &  gu$\rightleftarrows$en  & kk$\rightleftarrows$en &  ro$\rightleftarrows$en\\ \hline
        Foreign docs & 1.0m & 28k & 230k & 400k \\
      Paired docs & 745k & 7.3k & 80k & 270k \\
        First sents. & 205k & 3.2k & 52k & 78k\\
        Captions & 92k &  2.2k & 1.9k & 35k \\
        Comparable pairs & 0.1b & 14m & 32m & 64m \\
        Mined sents. & 1.7m & 49k & 183k & 675k\\
        BT & 2.1m & 1.5m & 2.2m & 2.1m\\
        Iterative BT & 4.0m & 3.8m & 4.0m & 6.1m\\
   \hline\hline
    \end{tabular}
   
    \caption{Data sizes for different pairs. We use a sample of English sentences with similar sizes to each data.} 
    \label{tab:par_sizes}

\end{table}

\paragraph{Pretraining} We pretrain four models on 3-tuples of languages via a single NVIDIA Geforce RTX 2080 TI with 11GB of memory. We create batches of 4K words, run pretraining for two million iterations where we alternate between language batches, and accumulate gradients for 8 steps. We use the apex library\footnote{\url{https://github.com/NVIDIA/apex}} to use FP-16 tensors. This whole process takes four weeks in a single GPU. We use the Adam optimizer~\cite{adam_kingma} with inverse square root and learning rate of $10^{-4}$, $4000$ warm-up steps, and dropout probability of $0.1$. 

\paragraph{Translation Training} Table~\ref{tab:par_sizes} shows the sizes of different types of datasets in our experiments. We pick comparable candidates for sentence pairs whose lengths are within a range of half to twice of each other. As we see, the final size of mined datasets heavily depends on the number of paired English-target language Wikipedia documents. We train our translation models initialized by pretrained models. More details about our hyper-parameters are in the supplementary material. All of our evaluations are conducted using SacreBLEU~\cite{post-2018-call} except for en$\leftrightarrow$ro in which we use BLEU score~\cite{papineni2002bleu} from Moses decoder scripts~\cite{koehn2007moses} for the sake of comparison to previous work.

\paragraph{Image Captioning} 
We use the Flickr~\cite{hodosh2013framing} and MS-Coco~\cite{chen2015microsoft} datasets for English\footnote{We have also tried Conceptual Captions~\cite{sharma2018conceptual} in our initial experiments but we have observed drops in performance. Previous work~\cite{singh2020we} have also observed a similar problem with Conceptual Captions as a noisy crawled caption dataset.}, and the gold-standard Arabic Flickr dataset~\cite{visapp20} for evaluation. The Arabic test set has $1000$ images with $3$ captions per image. We translate all the training datasets to Arabic for having translated caption data. The final training data contains $620K$ captions for about $125K$ unique images. Throughout experiments, we use the pretrained Resnet-152 models~\cite{resnet} from Pytorch~\cite{paszke2019pytorch}, and let it fine-tune during our training pipeline. Each training batch contains $20$ images. We accumulate gradients for $16$ steps, and use a dropout of $0.1$ for the projected image output representations. Other training parameters are the same as our translation training. To make our pipeline fully unsupervised, we use translated development sets to pick the best checkpoint during training.

\paragraph{Dependency Parsing} 
We use the Universal Dependencies v2.7 collection~\cite{universal2_7} for Arabic, Kazakh, and Romanian. We use the Stanza~\cite{stanza} pretrained supervised models for getting supervised parse trees for Arabic and Romanian, and use the UDPipe~\cite{straka2016udpipe} pretrained model for Kazakh. We translate about 2 million sentences from each language to English, and also 2 million English sentences to Arabic. We use a simple modification to Stanza to facilitate training on partially projected trees by masking dependency and label assignments for words with missing dependencies.  All of our training on projected dependencies is blindly conducted with $100k$ training steps with default parameters of Stanza~\cite{stanza}. As for gold-standard parallel data, we use our supervised translation training data for Romanian-English and Kazakh-English and use a sample of 2 million sentences from the UN Arabic-English data due to its large size that causes word alignment significant slow-down. For Kazakh \emph{wikily} projections, due to low supervised POS accuracy, we use the projected POS tags for projected words and supervised tags for unprojected words. We observe a two percent increase in performance by using projected tags.

\begin{table*}[t!]
    \centering \small
    \setlength{\tabcolsep}{5pt}
    \begin{tabular}{l l |c c c c c c c c}
    \hline\hline
        &  Model & ar$\rightarrow$en & en$\rightarrow$ar & gu$\rightarrow$en & en$\rightarrow$gu & kk$\rightarrow$en & en$\rightarrow$kk & ro$\rightarrow$en & en$\rightarrow$ro \\ \hline
     \parbox[t]{2mm}{\multirow{3}{*}{\rotatebox[origin=c]{90}{\tiny UNMT}}}
  &   \newcite{xlm_paper} & -- & -- & -- & -- &-- & --  & 31.8 & 33.3\\ 
         & \newcite{song2019mass} (MASS; 8 GPUs) & -- &-- &-- & -- &-- & --  & 33.1 & 35.2\\ 
      & Best published results & 11.0* & 9.4*  & $0.6^{1}$  & $0.6^{1}$ & $2.0^{1}$ & $0.8^{1}$ &$37.6^{4}$ & $36.3^{2}$ \\ \hline

      \parbox[t]{2mm}{\multirow{6}{*}{\rotatebox[origin=c]{90}{\tiny Wikily UNMT}}}   & First sentences + captions + titles  & 6.1 &  3.1 & 0.7  &  1.1 &   2.3 & 1.0  & 2.0 & 1.9\\
       &    Mined Corpora & 23.1  & 19.7  & 4.2  &  4.9 &  2.8  &  1.6  & 22.1 & 21.6  \\
       &    + Related Language  &  -- & --  &  9.1 &  7.8 & 7.3 &  2.3  &  23.2 &  21.5 \\
        &  + One-shot back-translation (bt-beam=4)  &  23.0 &  18.8 & {\bf 13.8} & 13.9 & 7.0 &  \colorbox{blue!20}{\bf 12.1} &  25.2 & 28.1 \\ 
        &     + Iterative back-translation (bt-beam=1)  & 24.4 &   18.9 &  13.3 &  \colorbox{blue!20}{\bf 15.2} &{\bf  9.0 }& 10.8  & 32.5 & 33.0 \\
        & + Retrain on mined data & {\bf 30.6} & {\bf 23.4} & -- & -- & -- & -- & -- & -- \\\hline
       &   (Semi-)Supervised & 48.9* &  40.6* & $14.2^{1}$ & $4.0^{1}$ & $12.5^{1}$ & $3.1^{1}$ & $39.9^{3}$ &  $38.5^{3}$ \\
      \hline\hline
    \end{tabular}
    \caption{BLEU scores for different models. Reference results are from *: Our implementation, 1:~\protect\newcite{kim-etal-2020-unsupervised}, 2:  \protect\newcite{Li2020Data-dependent}, 3: \protect\newcite{liu2020multilingual} (supervised), 4: \protect\newcite{tran2020cross} (unsupervised with mined parallel data).}
    \label{tab:main_results}
\end{table*}



\begin{table*}[]
     \centering \small
    \setlength{\tabcolsep}{3pt}
    \begin{tabular}{|l l | c| c | c c  c | c c  c | c c  c | }
    \hline 
     \multicolumn{2}{|c|}{\multirow{2}{*}{Method}} & \multirow{2}{*}{Version} & \multirow{2}{*}{Token and POS} & \multicolumn{3}{c|}{Arabic} & \multicolumn{3}{c|}{Kazakh} & \multicolumn{3}{c|}{Romanian} \\ \cline{5-13}
     & & & &    UAS & LAS&  BLEX & UAS & LAS & BLEX& UAS &LAS & BLEX \\ \hline \hline
   \parbox[t]{2mm}{\multirow{3}{*}{\rotatebox[origin=c]{90}{\tiny Previous}}}  & \newcite{rasooli-collins-2019-low} & 2.0 & gold/supervised & 61.2 & 48.8  & -- & -- & -- & -- & 76.3 & 64.3 & -- \\ 
      & \newcite{ahmad-etal-2019-difficulties} & 2.2 & gold & 38.1 & 28.0 & -- & -- & -- & -- & 65.1 & 54.1 & -- \\
     & \newcite{kurniawan2021ppt} & 2.2 & gold & 48.3 & 29.9 & -- & -- & -- & -- & -- & -- & -- \\ \hline
    \parbox[t]{2mm}{\multirow{4}{*}{\rotatebox[origin=c]{90}{\tiny Projection}}} &   \multirow{2}{*}{\emph{Wikily} translation} & \multirow{5}{*}{2.7}  & gold & 62.5 & 50.7 & 46.3 & 46.8 & 28.5 &  25.0 & 74.1 & 57.7 & 52.6 \\ 
      &  & &  supervised & 60.2 & 48.7 & 42.1 & 46.2 &  27.8 & 14.1 & 73.6 & 57.4 & 50.9 \\  \cline{2-2}\cline{4-13}
       &  \multirow{2}{*}{ Gold-standard Parallel data}  & & gold & 61.5 & 47.3 & 42.4 & 22.2 & 9.3 & 7.9 & 75.9 & 62.4 & 57.3 \\
    &    &   & supervised & 59.1 & 45.3 & 38.5 & 21.8 & 9.2 & 3.8 & 75.6 & 62.0 & 55.6    \\ \cline{1-2}\cline{4-13}
     \multicolumn{2}{|c|}{Supervised}   & & supervised & 84.2 & 79.8 & 72.7 & 48.0 & 29.8 & 13.7 & 90.8 & 86.0 & 80.0 \\ \hline \hline
    \end{tabular}
    \caption{Dependency parsing results on the Universal Dependencies dataset~\protect\cite{universal2_7}. Previous work has used different sub-versions of the Universal Dependencies data in which slight differences are expected.}
    \label{tab:projection_results}
\end{table*}

\begin{table}[t!]
     \centering \small
    \setlength{\tabcolsep}{3pt}
    \begin{tabular}{c c c c c c c c c c  }
    \hline 
       &    \multirow{2}{*}{Supervision}  & \multirow{2}{*}{Pretrained}  & \multicolumn{2}{c}{Multi-task}& \multicolumn{2}{c}{BLEU} \\ 
      &  & & EN & MT &  @1 & @4 \\ \hline \hline
  \parbox[t]{2mm}{\multirow{6}{*}{\rotatebox[origin=c]{90}{\tiny Translate train data}}} &  \emph{wikily}  & \xmark & \xmark & \xmark   & 33.1  &  4.57 \\ 
  &  \emph{wikily}  & \cmark & \xmark & \xmark   &  32.9 & 5.28  \\ 
    &  \emph{wikily}  & \cmark & \cmark & \xmark   &  32.8 & 4.37 \\
      &  \emph{wikily}  & \cmark & \xmark & \cmark   & 33.3 &  \colorbox{green!50}{5.72} \\ 
      &  \emph{wikily}  & \cmark & \cmark & \cmark   & \colorbox{green!50}{36.8} & 5.60 \\ 
      &  \emph{supervised}  & \cmark & \xmark & \xmark   & 17.7 & 1.26 \\      \hline
       \parbox[t]{2mm}{\multirow{4}{*}{\rotatebox[origin=c]{90}{\tiny Translate test}}} & 
      &   \multicolumn{3}{c}{English test performance$\rightarrow$}  & 68.7 & 20.42  \\ 
     &  \emph{wikily}  & \cmark &  \xmark &\xmark &  30.6 & 4.20 \\ 
       &  supervised  & \cmark &  \xmark& \xmark& 15.8 & 0.92\\ 
        &  Google  & \cmark & \xmark &\xmark & 31.8 & 5.56 \\ \hline
           & \multirow{2}{*}{Gold}  &  \cmark & \xmark & \xmark   &  33.7 & 3.76 \\
         &   &  \cmark & \cmark & \xmark   &  \colorbox{blue!30}{37.9} & 5.22  \\  \hline
    \end{tabular}
    \caption{Image captioning results evaluated on the Arabic Flickr dataset~\protect\cite{visapp20} using SacreBLEU~\protect\cite{post-2018-call}. ``pretrained'' indicates initializing our captioning model with our translation parameters.}
    \label{tab:caption_results}
\end{table}

\subsection{Translation Results}
Table~\ref{tab:main_results} shows the results of different settings in addition to baseline and state-of-the-art results. We see that Arabic as a clear exception needs more rounds of training: we train our Arabic model once again on mined data by initializing it by our back-translation model.\footnote{We have seen that during multi-tasking with image captioning, the translation BLEU score for Arabic-English significantly improves. We initially thought that multi-tasking is improving both translation and captioning, but our further investigation shows that it is actually due to lack of training for Arabic. We have tried the same procedure for other languages but have not observed any further gains.} We have not seen further improvement by back-translation. To have a fair comparison, we list the best supervised models for all language pairs (to the best of our knowledge). In low-resource settings, we outperform strong supervised models that are boosted by back-translation. In high-resource settings, our Arabic models achieve very high performance but regarding the fact that the parallel data for Arabic has 18M sentences, it is quite impossible to reach that level of accuracy.

\begin{figure*}[t!]
    \centering
    \includegraphics[width=0.95\textwidth]{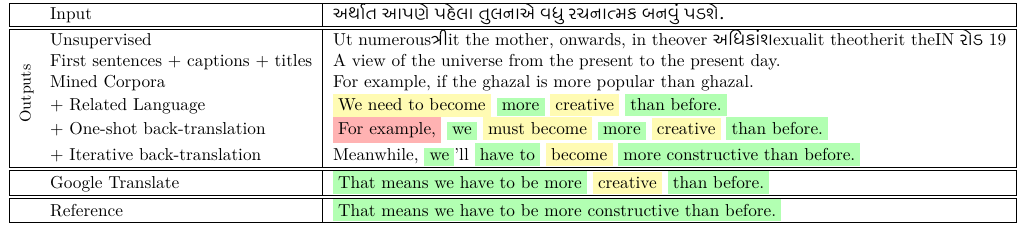}
    \caption{An example of a Gujarati sentence and its outputs from different models, as well as Google Translate.}
    \label{fig:output_example}
\end{figure*}

Figure~\ref{fig:output_example} shows a randomly chosen example from the Gujarati-English development data. As depicted, we see that the model after back-translation reaches to somewhat the core meaning of the sentence with a bit of divergence from exactly matching the reference. The final iterative back-translation output almost catches a correct translation. We also see that the use of the word ``creative'' is seen in Google Translate output, a model that is most likely trained on much larger parallel data than what is currently available for public use. In general, unsupervised translation performs very poorly compared to our approach in all directions.

\subsection{Captioning Results}
Table~\ref{tab:caption_results} shows the final results on the Arabic test set using the SacreBLEU measure~\cite{post-2018-call}. First, we should note that similar to \newcite{visapp20}, we see lower scales of BLEU scores due to morphological richness in Arabic. We see that if we initialize our model with the translation model and multi-task it with translation and also English captioning, we achieve much higher performance. It is interesting to observe that translating the English output on the test data to Arabic achieves a much lower result. 
This is a strong indicator of the strength of our approach. We also see that supervised translation fails to perform well. This might due to the UN translation training dataset which has a different domain from the caption dataset. Furthermore, we see that our model outperforms Google Translate which is a strong machine translation system, and that is actually what is being used as seed data for manual revision in the Arabic dataset.  Finally, it is interesting to see that our model outperforms supervised captioning. Multi-tasking make translation performance slightly worse.

\begin{figure*}[t!]
    \centering
        \setlength{\tabcolsep}{0pt}
    \begin{tabular}{c c}
       \includegraphics[width=0.36\textwidth]{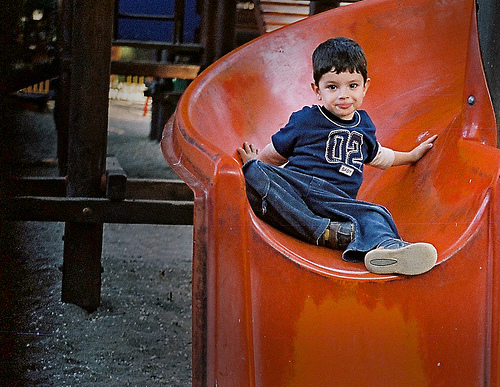} & \includegraphics[width=0.58\textwidth]{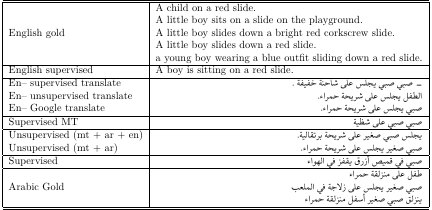} \\
    \end{tabular}
    \caption{An example of different outputs in our captioning experiments both for English and Arabic, as well as Arabic translations of English outputs on the Arabic Flickr dataset~\protect\cite{visapp20}.}
    \label{fig:caption_outputs}
\end{figure*}

Figure~\ref{fig:caption_outputs} shows a randomly picked example with different model outputs. We see that the two outputs from our approach with multi-tasking are roughly the same but one of them as more syntactic order overlap with the reference while both orders are correct in Arabic as a free-word order language. The word \setcode{utf8}
\<برتقالية >  means ``orange'' which is close to \setcode{utf8}
\<حمراء > that means ``red''. The word \setcode{utf8}
\<شريحة > means ``slide'' which is correct but other meanings of this word exist in the reference. In general, we observe that although superficially the BLEU scores for Arabic is low, it is mostly due to its lexical diversity, free-word order, and morphological complexity.

\subsection{Dependency Parsing Results}
Table~\ref{tab:projection_results} shows the results for dependency parsing experiments. We see that our model performs very high in Romanian with a UAS of $74$ which is much higher than that of \newcite{ahmad-etal-2019-difficulties} and slightly lower than that of \newcite{rasooli-collins-2019-low} which uses a combination of multi-source annotation projection and direct model transfer. Our work on Arabic outperforms all previous work and performs even better than using gold-standard parallel data. One clear highlight is our result in Kazakh. As mentioned before, by projecting the part-of-speech tags, we achieve roughly 2 percent absolute improvement. Our final results on Kazakh are significantly higher than that of using gold-standard parallel text (7K sentences).

\section{Related Work}
\newcite{kim-etal-2020-unsupervised} has shown that unsupervised translation models often fail to provide good translation systems for distant languages. Our work solves this problem by leveraging the Wikipedia data. Using pivot languages has been used in previous work~\cite{al-shedivat-parikh-2019-consistency}, as well as using related languages~\cite{zoph-etal-2016-transfer,nguyen-chiang-2017-transfer}. Our work only explores a simple idea of adding one similar language pair. Most likely, adding more language pairs and using ideas from recent work might improve the performance.

Wikipedia is an interesting dataset for solving NLP problems including machine translation~\cite{li2012wiki,patry-langlais-2011-identifying,lin2011unsupervised,tufics2013wikipedia,barron-cedeno-etal-2015-factory,wijaya2017learning,ruiter-etal-2019-self,srinivasan2021wit}. The WikiMatrix data~\cite{schwenk2019wikimatrix} is the most similar effort to ours in terms of using Wikipedia, but with using supervised translation models. Bitext mining has a longer history of research~\cite{resnik1998parallel,resnik2003web} in which most efforts are spent on using a seed supervised translation model~\cite{guo-etal-2018-effective,schwenk2019ccmatrix,artetxe-schwenk-2019-margin,schwenk2019wikimatrix,jones2021majority}. Recently, a number of papers have focused on unsupervised extraction of parallel data~\cite{ruiter-etal-2019-self,hangya-fraser-2019-unsupervised,keung2020unsupervised,tran2020cross,kuwanto2021lowresource}. \newcite{ruiter-etal-2019-self} focus on using vector similarity of sentences to extract parallel text from Wikipedia. Their work does not leverage structural signals from Wikipedia.  

Cross-lingual and unsupervised image captioning has been studied in previous work~\cite{gu2018unpaired,feng2019unsupervised,song2019unpaired,gu2019unpaired,gao2020unsupervised,burns2020learning}. Unlike previous work, we do not have a supervised translation model. Cross-lingual transfer of dependency parser have a long history. We encourage the reader to read a recent survey on this topic~\cite{das2020survey}. Our work does not use gold-standard parallel data or even supervised translation models to apply annotation projection.

\vspace{-.15cm}
\section{Conclusion}\label{sec:conclude}

We have described a fast and effective algorithm for learning translation systems using Wikipedia. We show that by wisely choosing what to use as seed data, we can have very good seed parallel data to mine more parallel text from Wikipedia. We have also shown that our translation models can be used in downstream cross-lingual natural language processing tasks. In the future, we plan to extend our approach beyond Wikipedia to other comparable datasets like the BBC World Service. A clear extension of this work is to try our approach on other cross-lingual tasks. Moreover, as many captions of the same images in Wikipedia are similar sentences and sometimes translations, multimodal machine translation  \cite{specia2016shared,caglayan2019probing,hewitt2018learning, yao2020multimodal} based on this data or the analysis of the data, such as whether more similar languages may share more similar captions \cite{khani2021cultural} are other interesting avenues.

\section*{Acknowledgments}
We would like to thank reviewers and the editor for their useful comments. We also would like to thank Alireza Zareian, Daniel (Joongwon) Kim, Qing Sun, and Afra Feyza Akyurek for their help and useful comments througout this project. This work is
supported in part by the DARPA HR001118S0044 (the LwLL program),
and the Department of the Air Force FA8750-19-
2-3334 (Semi-supervised Learning of Multimodal
Representations). The U.S. Government is authorized
to reproduce and distribute reprints for Governmental
purposes. The views and conclusions
contained in this publication are those of the authors
and should not be interpreted as representing
official policies or endorsements of DARPA, the
Air Force, and the U.S. Government.


\bibliography{refs}
\bibliographystyle{acl_natbib}
\appendix

\section{Cross-Lingual Embedding}
We use the off-the-shelf 300-dimensional FastText embeddings~\cite{grave-etal-2018-learning} as monolingual embedding vectors. We run FastAlign~\cite{dyer-etal-2013-simple} on the seed parallel text from both source-to-target and target-to-source directions, run alignment intersection to get intersected alignments, and extract the highest occurring alignment for every word as the dictionary entry. We make use of the cross-lingual CCA tool~\cite{faruqui-dyer-2014-improving} to extract 150-dimensional vectors. This tool can be run on a single CPU within a few hours.

\section{Monolingual and Translation Datasets}
We use an off-the-shelf Indic-transliteration library\footnote{\url{https://pypi.org/project/indic-transliteration}} to convert the Devanagari script to Hindi script to make the Hindi documents look like Gujarati by removing the graphical vertical bars from Hindi letters, thus increasing the chance of capturing more words in common. We boost the Romanian, Gujarati, and Kazakh monolingual data with newstext dataset from WMT.  For parallel data in similar languages, we use the Mizan parallel data for Persian~\cite{kashefi2018mizan} with one million sentences, the IITB data~\cite{kunchukuttan-etal-2018-iit} and HindiEnCorp 0.5~\cite{bojar2014hindencorp} for Hindi with a total of 367K sentences, ParaCrawl for Russian~\cite{espla-etal-2019-paracrawl} with 12M sentences, and Europarl for Italian~\cite{koehn2005europarl} with 2M sentences. We use the Arabic-English UN data~\cite{ziemski2016united}, WMT 2019 data~\cite{barrault-etal-2019-findings} for Gujarati-English and Kazakh-English, and WMT 2016 shared task data~\cite{bojar-etal-2016-findings} for Romanian-English. Following previous work~\cite{sennrich-etal-2016-edinburgh}, diacritics are removed from the Romanian data.

\section{Translation Training Parameters}

We pick comparable candidates for sentence pairs whose lengths are within a range of half to twice of each other. As we see, the final size of mined datasets heavily depends on the number of paired English-target language Wikipedia documents. We train our translation models initialized by pretrained models. Each batch has roughly 4K tokens. Except for Arabic, for which the size of mined data significantly outnumbers the size of Persian-English parallel data, we use the related language data before using iterative back-translation in which we only use the source and target monolingual datasets. We use similar learning hyper-parameters to pretraining except for iterative back-translation in which we accumulate gradients for 100 steps, and use a dropout probability of $0.2$ and $10000$ warmup steps since we find smaller dropout and warmup make the model diverge. Our one-shot back-translation experiments use a beam size of 4, but we use a beam size of one for iterative back-translation since we have not seen much gains in using beam-based iterative back-translation except for purely unsupervised settings. All of our translations are performed with a beam size of $4$ and $max\_len\_a=1.3$ and $max\_len\_b=5$. We alternate between supervised parallel data of a similar language paired with English and the mined data. 

We train translation models for roughly 400K batches except for Gujarati that has smaller mined data for which we train for 200K iterations. We have seen a quick divergence in Kazakh iterative back-translation, thereby we stopped it early after running it for one epoch of all monolingual data. Most likely, the mined data for Kazakh-English has lower quality (see the supplementary material for more details), and that leads to very noisy translations in back-translation outputs. All of our evaluations are conducted using SacreBLEU~\cite{post-2018-call} except for en$\leftrightarrow$ro in which we use BLEU score~\cite{papineni2002bleu} from Moses decoder scripts~\cite{koehn2007moses} for the sake of comparison to previous work.


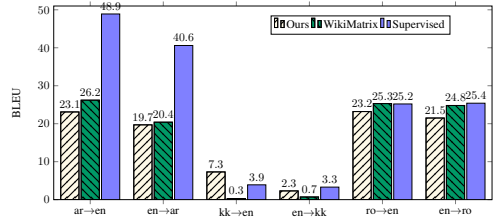
\begin{figure}[t!]
    \centering
    
\scalebox{0.4}{
\pgfplotstableread[row sep=\\,col sep=&]{
    direction & ours & wikimatrix & sup \\
   ar$\rightarrow$en & 23.1 &  26.2 & 48.9  \\
     en$\rightarrow$ar & 19.7 &    20.4 & 40.6
     \\  
     kk$\rightarrow$en & 7.3 &     0.3 & 3.9   \\ 
     en$\rightarrow$kk & 2.3 &    0.7 & 3.3   \\ 
     ro$\rightarrow$en & 23.2 &     25.3 & 25.2  \\ 
     en$\rightarrow$ro & 21.5 &     24.8 & 25.4   \\ 
    }\mydata

\begin{tikzpicture}
    \begin{axis}[
            ybar,
            bar width=.6cm,
            width=\textwidth,
            height=.5\textwidth,
            legend style={at={(0.7,0.95)},
                anchor=north,legend columns=-1},
            symbolic x coords={ar$\rightarrow$en,en$\rightarrow$ar,kk$\rightarrow$en,en$\rightarrow$kk,ro$\rightarrow$en,en$\rightarrow$ro},
            xtick=data,
            nodes near coords,
            nodes near coords align={vertical},
            ymin=0,ymax=51,
            ylabel={BLEU},
        ]
        \addplot[black,postaction={pattern=north east lines},fill=yellow!80!red!10!] table[x=direction,y=ours]{\mydata};
        \addplot[black,,postaction={pattern=north west lines},fill=green!60!blue] table[x=direction,y=wikimatrix]{\mydata};
        \addplot[black,fill=blue!50!] table[x=direction,y=sup]{\mydata};
        \legend{Ours, WikiMatrix, Supervised}
    \end{axis}
\end{tikzpicture}
}
\caption{Results using our mined data versus WikiMatrix~\protect\cite{schwenk2019wikimatrix} and gold-standard data.}
    \label{fig:compare_sup}
\end{figure}

\begin{figure}[t!]
    \centering
    
\scalebox{0.4}{
\pgfplotstableread[row sep=\\,col sep=&]{
    direction & nopret & withpret \\
   ar$\rightarrow$en & 18.6 & 23.0 \\
     en$\rightarrow$ar  & 14.9 & 19.7 \\  
      gu$\rightarrow$en  & 2.9 & 9.1\\ 
     en$\rightarrow$gu  & 3.6 & 7.8 \\ 
     kk$\rightarrow$en  & 2.4 & 7.3\\ 
     en$\rightarrow$kk  & 0.7 & 2.3\\ 
     ro$\rightarrow$en  & 19.5 & 23.2\\ 
     en$\rightarrow$ro  & 17.8 & 21.5 \\
    }\mydata

\begin{tikzpicture}
    \begin{axis}[
            ybar,
            bar width=.5cm,
            width=\textwidth,
            height=.5\textwidth,
            legend style={at={(0.5,0.9)},
                anchor=north,legend columns=-1},
            symbolic x coords={ar$\rightarrow$en,en$\rightarrow$ar,gu$\rightarrow$en,en$\rightarrow$gu,kk$\rightarrow$en,en$\rightarrow$kk,ro$\rightarrow$en,en$\rightarrow$ro},
            xtick=data,
            nodes near coords,
            nodes near coords align={vertical},
            ymin=0,ymax=27,
            ylabel={BLEU},
        ]
        \addplot[black,postaction={pattern=north east lines},fill=yellow!80!red!10!] table[x=direction,y=nopret]{\mydata};
        \addplot[black,fill=green!60!blue] table[x=direction,y=withpret]{\mydata};
        \legend{No Pretraining, With Pretraining}
    \end{axis}
\end{tikzpicture}
}
\caption{Results using mined data (no back-translation) with and without pretraining.}
    \label{fig:compare_pret}
\end{figure}
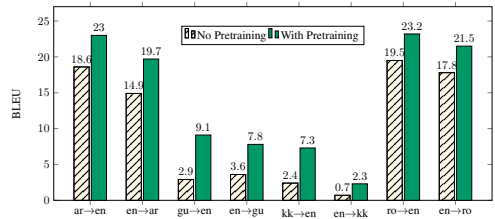
\begin{figure}[t!]
    \centering
    
\scalebox{0.4}{
\pgfplotstableread[row sep=\\,col sep=&]{
    direction & CRISS & Ours \\
   gu$\rightarrow$en & 18.0 & 13.8  \\
     en$\rightarrow$gu & 16.9 & 15.2 \\  
     kk$\rightarrow$en & 13.2 & 9.0  \\ 
     en$\rightarrow$kk & 4.3 & 10.8    \\ 
    }\mydata

\begin{tikzpicture}
    \begin{axis}[
            ybar,
            bar width=.6cm,
            width=\textwidth,
            height=.5\textwidth,
            legend style={at={(0.7,0.95)},
                anchor=north,legend columns=-1},
            symbolic x coords={gu$\rightarrow$en,en$\rightarrow$gu,kk$\rightarrow$en,en$\rightarrow$kk},
            xtick=data,
            nodes near coords,
            nodes near coords align={vertical},
            ymin=0,ymax=20,
            ylabel={BLEU},
        ]
        \addplot[black,postaction={pattern=north east lines},fill=blue] table[x=direction,y=CRISS]{\mydata};
        \addplot[black,,postaction={pattern=north west lines},fill=red] table[x=direction,y=Ours]{\mydata};
        \legend{CRISS, Ours}
    \end{axis}
\end{tikzpicture}
}
\caption{Our best results versus the \emph{supervised} model of \protect\newcite{tran2020cross}.}
    \label{fig:compare_tran}
\end{figure}
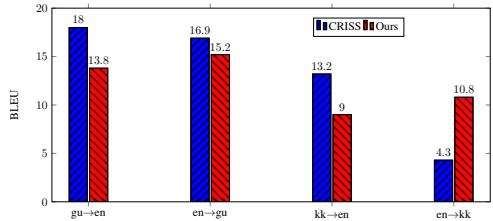



\section{Quality of Mined Data} The quality of parallel data matters a lot for getting high-accuracy. For example, we manually observe that the quality of mined data for all languages are very good except for Kazakh. Our hypothesis is that the Kazakh Wikipedia data is less aligned with the English content. We compare our mined data to that of the supervised mined data from WikiMatrix~\cite{schwenk2019wikimatrix} as well as gold-standard data. Figure~\ref{fig:compare_sup} shows the difference between the three datasets of three language pairs (WikiMatrix does not contain Gujarati). As we see, our data has BLEU scores near to WikiMatrix in all languages, and in the case of Kazakh, the model trained on our data performs higher than WikiMatrix. In other words, in the case of having very noisy comparable data, as is the case for Kazakh-English, our model even outperforms a contextualized supervised model. It is also interesting to see that our model outperforms the supervised model for Kazakh that has only 7.7K gold-standard training data. These are all strong evidences of the strength of our  approach in truly low-resource settings.

\section{Pretraining Matters} It is a truth universally acknowledged, that a single model in possession of a small training data and high learning capacity, must be in want of a pretrained model. To prove this, we run our translation experiments with and without pretraining. In this case, all models with the same training data and parameters are equal, but some models are more equal. Figure~\ref{fig:compare_pret} shows the results on the mined data. Clearly, there is a significant gain by using pre-trained models. For Gujarati, which is our the lowest-resource language in our experiments, the distance is more notable: from BLEU score of $2.9$ to $9.0$. If we had access to a cluster of high-memory GPUs, we could potentially obtain even higher results throughout all of our experiments. Therefore, we believe that part of the blame for  our results in English-Romanian is on pretraining. As we see in Figure~\ref{fig:compare_sup}, our supervised results without back-translation are also low for English-Romanian.  


\section{Comparing to CRISS} The recent work of \newcite{tran2020cross} shows impressive gains using high-quality pretrained models and iterative parallel data mining from a larger comparable data than that of Wikipedia. Their pretrained model is trained using  256 Nvidia V100 GPUs in approximately 2.5 weeks~\cite{liu2020multilingual}. Figure~\ref{fig:compare_tran} shows that by considering all these facts, our model still outperforms their \emph{supervised} model in English-to-Kazakh with a big margin ($4.3$ cs $10.8$) and gets close to their performance in other directions. We should emphasize on the fact that \newcite{tran2020cross} explores a much bigger comparable data than ours. One clear addition to our work is exploring parallel data from other available comparable datasets. Due to limited computational resources, we skip this part but we do believe that using our current unsupervised models can help extract even more high-quality parallel data from comparable datasets, and this might lead to further gains for low-resource languages.




\end{document}